\documentclass{SCIYA2023enOL}
\usepackage{booktabs}  
\usepackage{threeparttable} 
\usepackage{amsmath}
\usepackage{epstopdf}
\usepackage{multirow}

\makeatletter
\def\@fnsymbol#1{\ensuremath{\ifcase#1\or \dagger\or \ddagger\or
   \mathsection\or \mathparagraph\or \|\or * \dagger\dagger
   \or \ddagger\ddagger \else\@ctrerr\fi}}
\makeatother

\online
\begin{document}

\ensubject{fdsfd}
\ArticleType{ARTICLES}
\Year{2022}
\Month{January}
\Vol{65}
\BeginPage{1} %
\DOI{This paper has been accepted for publication in SCIENCE CHINA Mathematics}

\title[]{A Unified Pre-training and Adaptation Framework for Combinatorial Optimization on Graphs}{A Unified Pre-training and Adaptation Framework for Combinatorial Optimization on Graphs}

\author[1,$^\dagger$]{Ruibin Zeng}{zengruibin21@mails.ucas.ac.cn}
\author[2,\footnote{Equal contribution}]{Minglong Lei}{leiml@bjut.edu.cn}
\author[3,$^\dagger$]{Lingfeng Niu}{niulf@ucas.ac.cn}
\author[4,$\ast$]{Lan Cheng}{chenglan@hnfnu.edu.cn}

\AuthorMark{Ruibin Zeng, Minglong Lei}

\AuthorCitation{Ruibin Zeng, Minglong Lei, Lingfeng Niu, et al}

\address[1]{School of Computer Sciences, University of Chinese Academy of Sciences, Beijing, 100049, China}
\address[2]{Faculty of Information Technology, Beijing University of Technology, Beijing, 100124, China}
\address[3]{School of Economics and Management, University of Chinese Academy of Sciences, Beijing, 100190, China}
\address[4]{School of Mathematics and Statistics, Central South University, Changsha, 410075, China.}

\abstract{Combinatorial optimization (CO) on graphs is a classic topic that has been extensively studied across many scientific and industrial fields. Recently, solving CO problems on graphs through learning methods has attracted great attention. Advanced deep learning methods, e.g., graph neural networks (GNNs), have been used to effectively assist the process of solving COs. However, current frameworks based on GNNs are mainly designed for certain CO problems, thereby failing to consider their transferable and generalizable abilities among different COs on graphs. Moreover, simply using original graphs to model COs only captures the direct correlations among objects, which does not consider the mathematical logicality and properties of COs. In this paper, we propose a unified pre-training and adaptation framework for COs on graphs with the help of the maximum satisfiability (Max-SAT) problem. We first use Max-SAT to bridge different COs on graphs since they can be converted to Max-SAT problems represented by standard formulas and clauses with logical information. Then, we further design a pre-training and domain adaptation framework to extract the transferable and generalizable features so that different COs can benefit from them. In the pre-training stage, Max-SAT instances are generated to initialize the parameters of the model. In the fine-tuning stage, instances from CO and Max-SAT problems are used for adaptation so that the transferable ability can be further improved. Numerical experiments on several datasets show that features extracted by our framework exhibit superior transferability and Max-SAT can boost the ability to solve COs on graphs.}



\maketitle

\section{Introduction}

Combinatorial optimization (CO) is a multidisciplinary field ranging from optimization to artificial intelligence. The goal of CO is to find optimal solutions from finite possible options where exhaustive enumeration cannot be accomplished. Graphs are the most common research objects in CO. Typical CO problems on graphs include traveling salesman problems (TSP) \cite{pop2023comprehensive}, Max-Cut problems \cite{krislock2014improved}, and graph coloring problems \cite{wang2020reduction}. Solving CO on graphs can benefit many real-world applications, such as transportation logistics \cite{los1982combinatorial} and telecommunications \cite{martins2006metaheuristics}.

Solving COs on graphs is extremely difficult due to their discrete property and high computational costs. For many decades, researchers have made many efforts to develop various algorithms to solve COs on graphs. For example, branch and bound \cite{lawler1966branch} is a traditional exact algorithm that gradually divides the solution space into subspaces and uses bounds to prune possible solutions that exceed the bounds. Unfortunately, these traditional exact algorithms encounter excessive computational burdens as the scales of problems increase \cite{benson2000solving}. Recently, deep learning has developed as an effective tool for COs, inspiring the emergence of the category of learning-based methods for solving COs. Particularly, graph neural networks (GNNs) \cite{gilmer2017neural,xupowerful,kipfsemi} are advanced deep learning models designed for graph-structure data, which have been successfully applied to COs on graphs \cite{cappart2023combinatorial,khalil2017learning,li2018combinatorial}. In general, the graphs in CO problems are first fed into GNNs so that the high-level features can be extracted. Then, the predictions generated from features serve as feasible solutions or can be further used to guide the searching process of the heuristic learning process \cite{gasse2019exact,chen2022learning}. Learning-based methods extract more useful knowledge to reduce the search space, which can be used to effectively and efficiently solve COs on graphs. Besides, the data-driven manner is more adaptive to real-world problems when real data is available. 

However, current GNN models for CO problems still face several challenges.
First, directly modeling CO with graphs is intuitive but not necessarily accurate. Although graphs can capture relationships between objects, they ignore the mathematical logicality and properties inside COs. Therefore, constructing learning methods over original graphs often results in insufficient problem-solving abilities. Second, current methods are usually designed to solve a single CO problem on graphs. Nevertheless, different COs on graphs may share the same graphs and leverage similar GNN backbones for feature extraction. It is quite beneficial to explore whether the transferability of knowledge exists among different problems. Moreover, for real-world problems on graphs, the data collected for training GNNs is often limited, it is essential to leverage data from other problems that may have common knowledge to improve the generalization ability of GNNs.

To address the above issues, we propose in this paper a unified pre-training and adaptation framework for COs on graphs. The core idea is to develop a generalizable and transferable GNN framework from which all COs can be benefited. To achieve this goal, it is necessary to find out the common knowledge and properties of all CO problems on graphs that can be extracted by GNNs and used for different problems. We notice that using graphs to formulate CO omits logical information that is usually more general and common for all problems. Inspired by this observation, we propose to leverage the maximum satisfiability problems (Max-SAT) to bridge different CO problems on graphs. Max-SAT is a classic problem containing propositional logic formulas that can be used to describe additional logic information beyond simple relations on graphs. With this unified representation, we further adopt a pre-training process and a fine-tuning process based on domain adaptation to extract general features and utilize the transferability between different problems. Our framework can be adaptive to all CO problems that can be transferred to Max-SAT problems and is also suitable for various GNN architectures, which exhibit superior flexibility and versatility.

To be concrete, we first use the Max-SAT problem as an intermediate problem by which we can transform graphs from different COs to clauses and formulas. We can construct bipartite graphs from these clauses in Max-SAT that not only capture logical information but also can be adaptive to different COs on graphs. In the training stage, we can generate abundant clauses from certain Max-SAT problems without considering the original CO and transform them into bipartite graphs for pre-training. Due to the generic nature of Max-SAT, the pre-trained model is equipped with better initialization. Then, we use the original CO and Max-SAT to construct an adaptation framework with adversarial training for fine-tuning, which can further utilize the transferable information in Max-SAT. In both the pre-training and fine-tuning stages, the graph attention networks are leveraged as backbones to extract information in bipartite graphs transformed from COs on graphs and Max-SAT. The bipartite graph attention networks separately consider the aggregation of clauses and variables, which can better capture the dependencies between variables and clauses with different neighbor aggregation processes. 

To summarize, the contributions of this work are as follows:
\begin{itemize}
\item [(1)]
   We propose a unified pre-training and adaptation framework based on Max-SAT that can learn generalizable and transferable information to benefit the solving of COs on graphs. 
  \item [(2)]
  We propose to use Max-SAT to bridge various COs on graphs, by which all problems can be represented as a unified form. The Max-SAT instances also carry additional logical relationships that can be further utilized by GNNs.
  \item [(3)]
  We design a pre-training and domain adaptation architecture to extract generalizable and transferable features based on instances generated from the Max-SAT and COs on graphs. This framework is versatile to various COs and can be used to boost the ability to solve these problems.
  \item [(4)]
  We evaluate our method on both synthetic datasets and open benchmarks. The numerical results demonstrated the effectiveness of our framework in solving COs on graphs.
\end{itemize}

The remainder of this paper is organized as follows. Section 2 presents the learning methods for COs and popular GNN architectures for feature extraction on graphs. Section 3 introduces the problem transfer, pre-training, and fine-tuning procedures in our framework and summarizes the overall algorithms for training. The experiments that aim to address our three main claims are listed in Section 4. We conclude our work in Section 5.  

\section{Related Work}
\subsection{Learning Methods for Solving CO Problems}
Traditional methods for solving CO mainly include branch and bound, dynamic programming, and heuristic algorithms. Branch and bound is a branch-and-prune search method used to reduce the search space \cite{fischetti1994branch, pardalos1992branch, lawler1966branch}. It progressively decomposes the problem into smaller subproblems and applies pruning strategies to exclude branches that cannot produce the optimal solution. Dynamic programming breaks down a large problem into smaller subproblems and solves them by utilizing the relationships between the subproblems \cite{bellman1966dynamic, held1962dynamic, martello1999dynamic, wolsey1999integer}. Dynamic programming is suitable for problems with overlapping subproblem structures, such as the knapsack problem \cite{bertsimas2002approximate} and TSP \cite{bouman2018dynamic}. Heuristic methods employ heuristic strategies or random operations to search the solution space and continuously improve the current solution until a satisfactory solution is found or a predetermined stopping condition is met. Common heuristic algorithms include greedy algorithms \cite{yagiura2001metaheuristic}, local search \cite{orlin2004approximate}, simulated annealing \cite{aarts1988quantitative}, and genetic algorithms \cite{berger2003hybrid}. Traditional methods for solving CO problems have limitations including high computational complexity, lack of scalability for large-size problems, and limited solution quality in terms of finding globally optimal solutions. 

In recent years, deep learning methods have emerged as promising approaches to tackle these problems by leveraging the power of data-driven modeling and computational intelligence. One category of methods approximates the process of solving CO problems through deep learning methods \cite{chassein2020approximating}. By learning from a dataset of problem instances, these deep learning models can capture intrinsic patterns and dependencies in the problem and the predictions are then utilized to guide the search process toward better solutions. These approaches offer advantages in terms of speed and scalability as they reduce the requirements of expensive evaluations for the objective function during the search. Another category of methods is based on learning-based heuristics and meta-heuristics \cite{khalil2017learning, lemos2019graph, li2018combinatorial, gasse2019exact}. Deep learning models can be employed to create intelligent decision rules or policies to guide the search process. Reinforcement learning techniques have demonstrated success in learning effective exploration-exploitation strategies for CO problems \cite{kim2022sym,kool2018attention}. By incorporating learning-based heuristics into traditional optimization algorithms, better-quality solutions can be obtained with reduced computational effort.

\subsection{Graph Neural Networks}
GNNs belong to a typical deep-learning framework for graphs that follow a message-passing mechanism \cite{gilmer2017neural}. Graph convolution networks (GCNs) \cite{kipfsemi} define graph convolutions on the spectral domain so that the structural information from neighbors can be aggregated. GraphSAGE \cite{hamilton2017inductive} proposes an inductive representation learning method based on aggregation and sampling directly on the spatial domain. Graph attention networks (GATs) \cite{velivckovic2018graph} utilize self-attention mechanisms to learn node features where the importance of neighbors can be considered during the aggregation of information. More recently, prompt tuning has been used to generalize GNNs \cite{sun2022gppt, yi2023contrastive}. GPPT \cite{sun2022gppt} proposes a graph prompting function for GNNs so that the pre-trained models can predict labels without fine-tuning. Since GNNs have been successfully used in various downstream tasks, researchers pay attention to the theories behind GNNs. For example, over-smoothing \cite{bodnar2022neural} and over-squashing \cite{di2023over} are the two main issues for GNNs. BORF \cite{nguyen2023revisiting} proposes a rewiring algorithm based on Ollivier-Ricci curvature to relieve both over-smoothing and over-squashing problems for GNNs. The expressive abilities of GNNs have also been studied to demonstrate why GNNs perform well in many tasks \cite{xupowerful, zhang2022rethinking, agarwal2023evaluating}.

Recently, GNNs have been used in graph domain adaptation to address the domain shift on graphs \cite{yuan2022self}. Domain adaptation on graphs aims at leveraging knowledge learned from the source domain to improve the performance of a GNN model on a different target domain. Adversarial training is often used to build the alignment framework \cite{dai2022graph, zhu2023cross}. For example, TFLG \cite{zhu2023cross} utilizes instance and class levels of structures and leverages adversarial training to learn domain-invariant features. Except for the features that can be easily used for domain alignment, many researchers seek to learn invariant information from different levels. GCAN \cite{ma2019gcan} is a graph convolutional adversarial network introduced for unsupervised domain adaptation by jointly modeling structures, domain labels, and class labels. StruRW \cite{liu2023structural} presents a structural re-weighting method to address the conditional structure shift problem. Since spectral properties play an important role in graph structures, many works also attempt to explore the alignment of spectral information \cite{pil2023domain, you2023graph}.

\section{Methodology}
\subsection{CO Problems on Graphs}

Given a graph $\mathcal{G}=(\mathcal{V}, \mathcal{E})$, where $\mathcal{V}$ is a set of nodes and $\mathcal{E}$ is a set of edges between nodes. The CO problems on graphs are to find a subset of nodes $\mathcal{S}$ under given constraint conditions so that the number of edges between the nodes in $\mathcal{S}$ satisfies a specific condition and the objective function can be maximized or minimized.
Generally, the CO can be expressed as:
\begin{subequations}
\begin{align}
\centering
&\mathop{\max} f(\mathcal{S}), \\
&g_i(\mathcal{S})\leq  b_i, b_i\in \Omega,
\end{align}
\end{subequations}
where $f(\mathcal{S})$ is the objective function, $g_i(\mathcal{S})$ is the constraint condition function, $b_i$ is the boundary of the constraint condition, and $\Omega$ is the set of constraint indicators. In the following part, we give three concrete CO problems on graphs.

\paragraph{Max-Cut Problem}
Given a graph $\mathcal{G}=(\mathcal{V}, \mathcal{E})$, the goal of the Max-Cut problem is to find a partition that divides $\mathcal{V}$ into subset $\mathcal{S}$ and its complementary set $\mathcal{S}^{\prime}=\mathcal{V}-\mathcal{S}$ where the number of cuts between $\mathcal{S}$ and $\mathcal{S}^{\prime}$ is maximized. The cuts refer to the edges between $\mathcal{S}$ and $\mathcal{S}^{\prime}$.
The objective function of Max-Cut can be denoted as:
\begin{equation}
f(\mathcal{S})=\sum\nolimits_{e_{i,j}\in \mathcal{E}} (v_i+v_j-2v_iv_j),
\end{equation}
where $v_i=1$ when $v_i\in \mathcal{S}$ and $v_i=0$ otherwise, and $e_{i,j}$ is the edge between nodes $v_i$ and $v_j$.

\paragraph{Maximum Independent Set (MIS) Problem}
Given a graph $\mathcal{G}=(\mathcal{V}, \mathcal{E})$, the maximum independent set problem aims to find a node set $\mathcal{S}$ with a maximal number of nodes where there are no edges between any two nodes in this set.
The objective function and constraint conditions of the maximum independent set problem can be denoted as:
\begin{subequations}
\begin{align}
&f(\mathcal{S})=\sum\nolimits_{v_i\in \mathcal{V}} v_i,\\
&v_i+v_j\leq 1\lvert \forall e_{i,j}\in \mathcal{E},
\end{align}
\end{subequations}
where $v_i=1$ when $v_i\in \mathcal{S}$ and $v_i=0$ otherwise.

\paragraph{Minimum Dominated Set (MDS) Problem}
Given a graph $\mathcal{G}=(\mathcal{V}, \mathcal{E})$, the goal of the minimum dominated set problem is to find a dominated set $\mathcal{S}$ where each node $v_i\in \mathcal{V}$ or at least one of its neighbor nodes belongs to the dominated set $\mathcal{S}$. The number of nodes in $\mathcal{S}$ should also be minimized.
The objective function and constraint conditions of the minimum dominated set problem can be denoted as:
\begin{subequations}
\begin{align}
\centering
&f(\mathcal{S})=-\sum\nolimits_{v_i\in \mathcal{V}} v_i,\\
&v_i+v_j\geq 1\lvert e_{i,j}\in \mathcal{E},\forall v_i\in \mathcal{V},
\end{align}
\end{subequations}
where $v_i=1$ when $v_i\in \mathcal{S}$ and $v_i=0$ otherwise.

\subsection{Problem Transfer via Max-SAT Problem}\label{tranfer_G}
In this subsection, we introduce the details of leveraging the Max-SAT problem to bridge different CO problems. In general, to fully utilize the common logical information in COs, we first convert the original graphs of the CO problems into several clauses of Max-SAT problems and then transform these clauses into bipartite graphs.

\subsubsection{From Graph to Clauses}
Given a graph from any CO, we use clauses with variables to describe the objective function and the constraint conditions.
Max-SAT is an optimization problem that contains a set of clauses and variables. Define $m$ clauses $\mathcal{C}= \{c_1,c_2,\cdots,c_m\}$ and $n$ variables $\mathcal{X}=\{x_1,x_2,\cdots,x_n\}$, the Max-SAT problem is to find a truth assignment of variables so that the number of satisfied clauses is maximized. 

To generate clauses, we define two rules for the objective function and constraint conditions respectively. The soft rule is based on the objective function and the hard rule is based on the constraint conditions. The difference between those two rules mainly lies in whether the clauses should be strictly satisfied. For clarification, we let the variable $x_i$ correspond to the node $v_i$ in graph $\mathcal{G}$. For demonstration purposes, we provide the details of transformations for Max-Cut, MIS, and MDS problems.

For every $v_i\in \mathcal{V}$ in MIS problems, the clauses are generated by:
\begin{subequations}
\begin{align}
\centering
&c_h=\neg x_{i}\vee \neg x_{j}|e_{i,j}\in \mathcal{E}, \\
&c_s= x_{i}|v_{i} \in \mathcal{V},
\end{align}
\end{subequations}
where $c_h \in \mathcal{C}_{h}$ denotes the hard clause which describes that the set cannot contain the node and its neighbors at the same time, $c_s \in \mathcal{C}_{s}$ denotes the soft clause which describes that the number of nodes in the MIS problem should be maximized, $\neg$ and $\vee$ are the negation and disjunction.

For every $v_i\in \mathcal{V}$ in MDS problems, the clauses are generated by:
\begin{subequations}
\begin{align}
\centering
&c_h=x_{i}\vee x_{j}|e_{i,j}\in \mathcal{E}, \\
&c_s= \neg x_{i}|v_{i} \in \mathcal{V},
\end{align}
\end{subequations}
where $c_h \in \mathcal{C}_{h}$ describes that the node or at least one of its neighbor nodes belongs to MDS, and $c_s \in \mathcal{C}_{s}$ describes that the number of nodes in MDS should be minimized.

For every $e_{i,j}\in \mathcal{E}$ in Max-Cut problems, the clauses are generated by:
\begin{subequations}
\begin{align}
\centering
&c_s=x_{i}\vee x_{j}, \\
&c_{s}^{\prime}=\neg x_{i}\vee \neg x_{j},
\end{align}
\end{subequations}
where $c_s$ and $c_{s}^{\prime}$ are all soft clauses belonging to $\mathcal{C}_{s}$ that describe the number of cut edges that should be maximized.

\subsubsection{From Clauses to Bipartite Graph}

The soft and hard clauses describe the relations between clauses and variables, which reveals the logical dependencies. Then, we can use bipartite graphs to further represent these relations in Max-SAT. In general, the bipartite graph is denoted as $\mathcal{\tilde{G}}=(\mathcal{\tilde{V}}_x,\mathcal{\tilde{V}}_c,\mathcal{\tilde{E}})$ where $\mathcal{\tilde{V}}_x$ and $\mathcal{\tilde{V}}_c$ are node sets, and $\mathcal{\tilde{E}}$ is the edge set. The main difference between graphs and bipartite graphs is that the nodes in bipartite graphs belong to two different types. The variables $\mathcal{X}$ correspond to the set of nodes $\mathcal{\tilde V}_x$ and the clauses $\mathcal{C}$ correspond to the set of nodes $\mathcal{\tilde V}_c$. To generate the bipartite graph, the core step is to construct the edges between nodes. For each variable $x_i$ in each clause $c_i$, an edge $\tilde e_{i,j}\in\mathcal{\tilde E}$ corresponds to the affiliation between the variable and the clause.

\begin{figure}[h]
         \centering
         \includegraphics[width=0.7\linewidth]{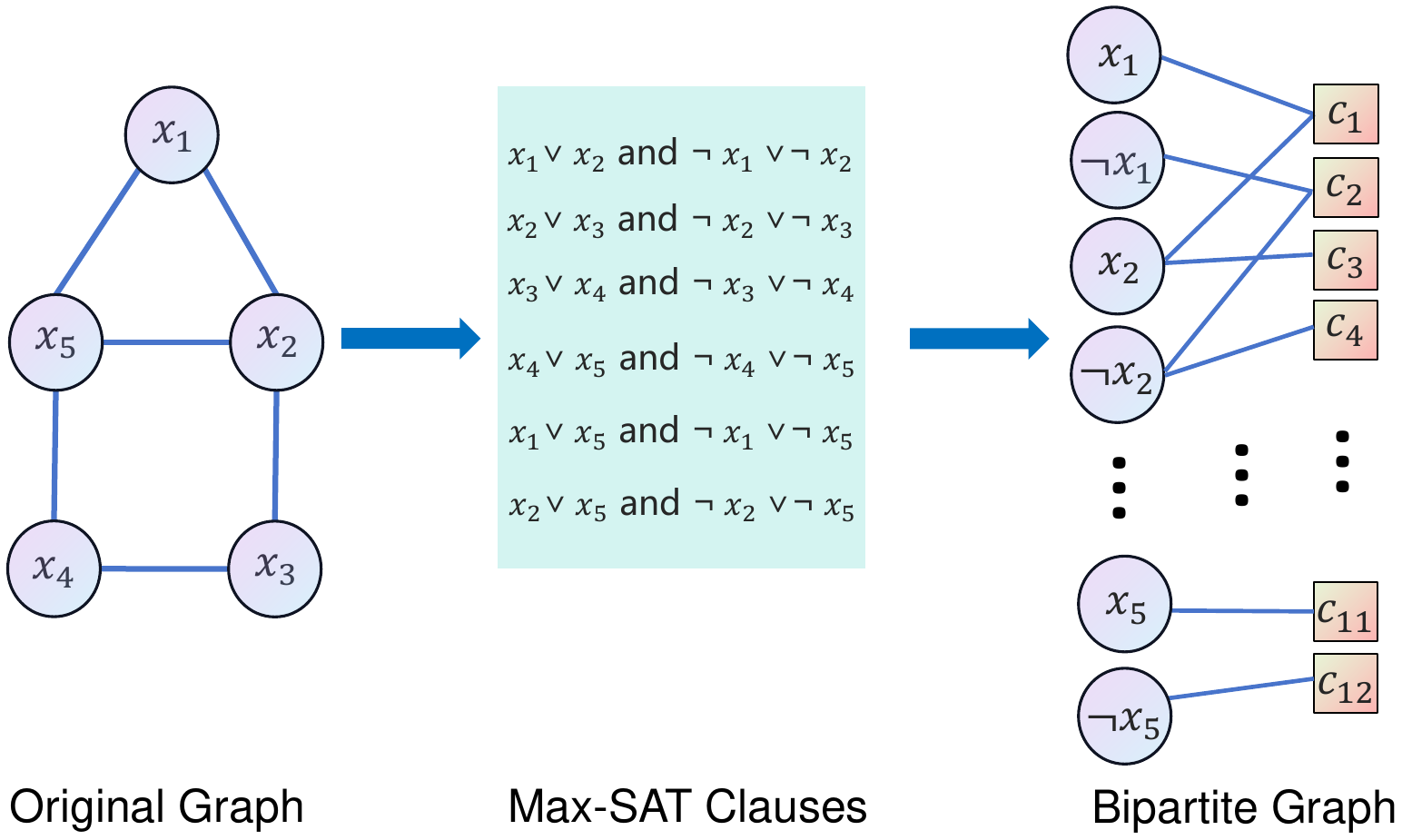}
         \caption{Problem transfer process of our framework via Max-SAT. The input graph of CO problems is first transferred as Max-SAT clauses. Then, a bipartite graph is generated from these clauses by the affiliations between variables and clauses.}
         \label{problem transfer}
\end{figure}

In this transfer process, we transfer the origin graph $\mathcal{G}=(\mathcal{V}, \mathcal{E})$ of the CO problem into the bipartite graph $\mathcal{\tilde{G}}=(\mathcal{\tilde{V}}_x,\mathcal{\tilde{V}}_c, \mathcal{\tilde{E}})$. A concrete transfer process can be shown in Figure \ref{problem transfer}. We choose Max-Cut as an example for demonstration. The first step is to transfer the graph in Max-Cut into a set of clauses in Max-SAT. We use the hard and soft rules to generate these clauses. After generating these clauses, we then transfer them into a bipartite graph. The variables and clauses are set as nodes and the affiliations between variables and clauses constitute edges. The new bipartite graph can be then used for the next feature extraction process.

Notice that transferring graphs in original COs to bipartite graphs has two main advantages: First, original graphs only contain simple dependencies between variables. In contrast, the bipartite graphs transferred from Max-SAT clauses are then equipped with logical information from COs; Second, the bipartite graphs are general forms that can represent various COs on graphs since all of these COs can be transformed into Max-SAT problems. Therefore, we can easily construct a knowledge transfer framework to leverage the generalizable and transferable information of these problems with the help of bipartite graphs. 

\subsection{Learning with Bipartite Graph Attention Networks}\label{GAT}

In this subsection, we first introduce the backbone network that is used to extract features from bipartite graphs. Traditional GNNs for feature extractions from graphs mainly follow a message-passing scheme:
\begin{align}
    \mathbf{a}_{v}^{(l)} = &\operatorname{AGG}(\mathbf{x}_{u} \mid u \in \mathcal{N}(v)) \label{agg},\\
    \mathbf{x}_{v}^{(l+1)} = &\operatorname{COM}(\mathbf{a}_{v}^{(l)},\mathbf{x}_{v}^{(l)}),
    \label{update}
\end{align}
where $u$ and $v$ denote the nodes, $\mathbf{a}_{v}$ denotes the feature vector accumulated from neighbor nodes, $\mathbf{x}_{v}^{(l)}$ denotes the features at layer $l$ and $\mathbf{x}_{v}^{(0)}$ is the initial attributes, $\operatorname{AGG}(\cdot)$ and $\operatorname{COM}(\cdot)$ are the aggregation and combination functions for message passing and feature updating, and $\mathcal{N}(v)$ denotes the set of neighbors for $v$.

The above process is only suitable for graphs with one type of nodes. Recall that a bipartite graph contains two sets of nodes that represent variables and clauses and a set of edges that connect nodes from these two sets. Due to the different properties of nodes, the traditional message-passing scheme in GNNs cannot be directly used to extract features. We particularly design a bipartite GNN with attentive message-passing schemes.

For an input bipartite graph $\mathcal{\tilde{G}}=(\mathcal{\tilde{V}}_x,\mathcal{\tilde{V}}_c, \mathcal{\tilde{E}})$, the attributed matrices of variables and clauses are denoted as $\mathbf{C} \in  \mathbb{R}^{m \times d}$ and $\mathbf{X} \in \mathbb{R}^{n \times d}$ where $m$ and $n$ denote the numbers of clauses and variables, and $d$ is the dimension of attributes. For symbolic convenience, we use $v_{x}(i)$ and $v_{c}(i)$ to denote nodes of variables and clauses in the bipartite graph. The variable and clause of these nodes are denoted as $x_i$ and $c_i$ and their attributed vectors are denoted as $\mathbf{x}_{i}$ and $\mathbf{c}_{i}$ accordingly. 

Since the initial bipartite graph does not contain attributed information, the attributed matrices require initialization. There are different strategies for initialization, e.g., uniform distribution, normal distribution, and all-one matrices. Since no significant influences have been observed in our experiments, we simply initialize attributed matrices as all-one matrices $\mathbf{X}_{ini}$ and $\mathbf{C}_{ini}$.

The message-passing process of bipartite GNNs consists of two steps in each iteration, the \textit{clause-wise aggregation} step and the \textit{variable-wise aggregation} step. First, clause-wise aggregation updates the feature of the clause node by aggregating the features from the variable nodes. To further discriminate the importance of different neighbors, the attention mechanism is introduced to learn features during the aggregation. Given a clause $c_i$ and its neighbor variable $x_j$, the layer-wise aggregation from $x_j$ to $c_i$ through attention is represented as:
\begin{align}\label{var_to_cla}
{\alpha}_{x_j \rightarrow c_i}^{(l)} =&\frac{\exp \Big(\langle{(\textbf{w}_{Q}^{(l)}{\textbf{c}_i^{(l)}}),(\textbf{w}_{K}^{(l)}{\textbf{x}_j^{(l)}}})\rangle \Big)}{\sum_{v_x(k) \in \mathcal{N}(v_{c}(i))} \exp\Big(\langle{(\textbf{w}_{Q}^{(l)}{\textbf{c}_i^{(l)}}),(\textbf{w}_{K}^{(l)}{\textbf{x}_k^{(l)}}})\rangle\Big)}, \\ 
\mathbf{c}_{i}^{(l+1)} = &\operatorname{MLP}\Big({\textbf{c}_i^{(l)}},\sum_{v_x(j) \in \mathcal{N}(v_{c}(i))}{\alpha}_{x_j \rightarrow c_i}^{(l)}(\textbf{w}_{V}^{(l)}\textbf{x}_{j}^{(l)})\Big),
\label{var_to_cla2}
\end{align}
where $\alpha_{x_j \rightarrow c_i}$ denotes the attention score between clause $c_i$ and variable $x_j$, $\textbf{w}_{Q}^{(l)}\in \mathbb{R}^{d}$, $\textbf{w}_{K}^{(l)}\in \mathbb{R}^{d}$, and $\textbf{w}_{V}^{(l)}\in \mathbb{R}^{d}$ are learnable parameters of weights, $\textbf{c}_i^{(l)}$ and $\textbf{x}_j^{(l)}$ are features of $i$-th clause and $j$-th variable at layer $l$, $\mathcal{N}(v_c(i))$ is the set of neighbor variable nodes for clause $c_{i}$, $\langle \cdot, \cdot \rangle$ is the dot product operation, $\textbf{c}_{i}^{(l+1)}$ is the updated features for clause $c_i$, and MLP denotes the multi-layer perceptron.

Second, the variable feature can be updated through variable-wise aggregation from the features of clauses that contain the variable. Similarly, for variable $x_i$, the layer-wise aggregation from $c_j$  to $x_i$ can be denoted as:
\begin{align}\label{cla_to_var}
{\alpha}_{c_j \rightarrow x_i}^{(l)}=&\frac{\exp \Big(\langle{(\tilde{\textbf{w}}_{Q}^{(l)}{\textbf{x}_i^{(l)}}),(\tilde{\textbf{w}}_{K}^{(l)}{\textbf{c}_j^{(l)}}})\rangle \Big)}{\sum_{v_c(k) \in \mathcal{N}(v_{x}(i))} \exp\Big(\langle{(\tilde{\textbf{w}}_{Q}^{(l)}{\textbf{x}_i^{(l)}}),(\tilde{\textbf{w}}_{K}^{(l)}{\textbf{c}_k^{(l)}}})\rangle\Big)}, \\
\mathbf{x}_{i}^{(l+1)}= &\operatorname{MLP}\Big({\textbf{x}_i^{(l)}},\sum_{v_c(j) \in \mathcal{N}(v_{x}(i))}{\alpha}_{c_j \rightarrow x_i}^{(l)}(\tilde{\textbf{w}}_{V}^{(l)}\textbf{c}_j^{(l)})\Big),
\label{cla_to_var2}
\end{align}
where $\alpha_{c_j \rightarrow x_i}$ denotes the attention score between clause $c_j$ and variable $x_i$, $\tilde{\textbf{w}}_{Q}^{(l)}$, $\tilde{\textbf{w}}_{K}^{(l)}$, and $\tilde{\textbf{w}}_{V}^{(l)}$ are parameters, and $\mathcal{N}(v_x(i))$ is the neighbor set of clause nodes for variable $x_{i}$.

After the two steps of aggregation, we can obtain the updated features of variables $\mathbf{X}^{(L)}$ and clauses $\mathbf{C}^{(L)}$ where $L$ is the number of layers for bipartite GNNs. The correlations between variables and clauses can be captured by the aggregation processes and the importance of neighbors can be learned by the attention mechanism. We can then build our framework based on the bipartite GNN backbone.

\begin{figure*}
         \centering
         \includegraphics[width=\linewidth]{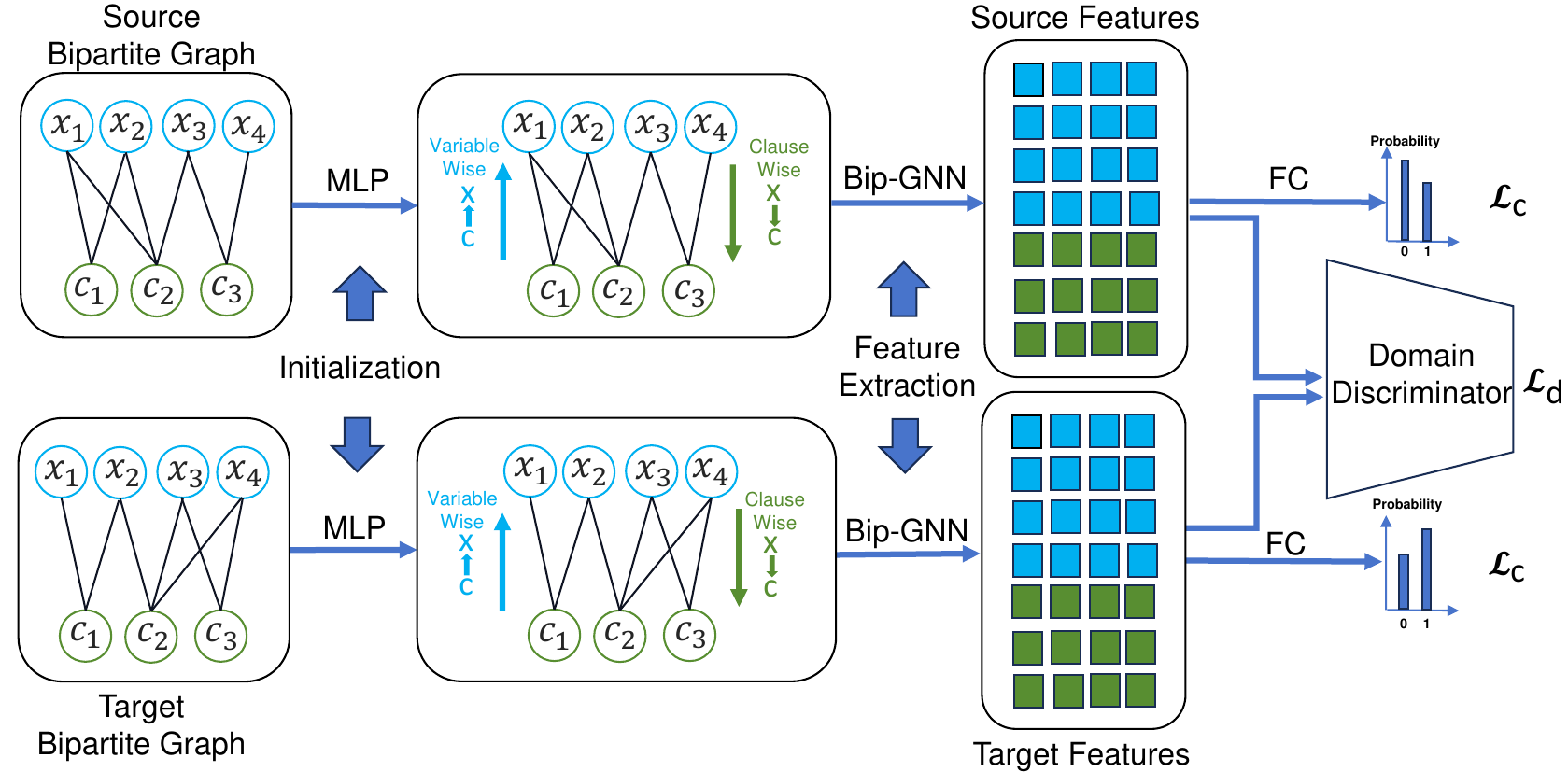}
         \caption{Illustration of the proposed pre-training and fine-tuning architecture. In the pre-training stage, the bipartite graphs generated from Max-SAT clauses are used to train the MLP, bipartite GNN backbone ($\operatorname{Bip-GNN(\cdot)}$), and fully connected classification layers $\operatorname{FC}(\cdot)$. In the fine-tuning stage, the bipartite graphs generated by Max-SAT and the CO are treated as source and target domains and are separately fed into the MLP and $\operatorname{Bip-GNN}$ to obtain features. A discriminator is to obtain domain labels for domain adaptation. The features are passed through the classification layer to predict the labels of variable nodes.}
         \label{pretrain_and_finetune}
\end{figure*}

\subsection{Pre-Training and Fine-Tuning with Domain Adaptation}

In this subsection, we introduce the framework of our work for training as depicted in Figure \ref{pretrain_and_finetune}. Our framework consists of two stages. The first stage is to pre-train our model by using massive samples generated from Max-SAT and the second stage is to further fine-tune the model with domain adaptation that combines samples from Max-SAT and CO problems.

\subsubsection{Pre-Training}
In the pre-training stage, the goal is to learn the general knowledge from Max-SAT and obtain better parameter initialization that can be used for solving different CO problems. We first generate massive clauses from Max-SAT and then convert them into bipartite graphs.

To generate Max-SAT clauses, we follow \cite{2009satindustry} and select different distributions to simulate different scenarios of problems: (1) In uniform distribution, the variables and clauses appear with the same frequency. The sizes of clauses also appear with the same frequency. The clause size refers to the number of variables within a clause. For example, given a clause $x_1 \vee  x_2 \vee  \neg x_3$, the clause size is three. (2) The single power-low distribution is a non-uniform distribution where most variables appear in low frequencies while only a few variables appear in high frequencies. The sizes of clauses also appear with the same frequency. (3) Double power-low distribution is also a non-uniform distribution that aims to simulate extremely uneven samples. Small clause size appears in high frequencies and the frequency decreases as the clause size increases.

After obtaining the bipartite graphs, we can build our pre-training process. To be concrete, the model structure for pre-training contains three parts, an MLP that is used to map the initial node attributes
$\textbf{X}_{ini}$ and $\textbf{C}_{ini}$ into low-dimensional features $\textbf{X}^{(0)}$ and $\textbf{C}^{(0)}$; the bipartite GNN backbone $\operatorname{Bip-GNN(\cdot)}$ from section \ref{GAT} extracts high-level node features for variables $\textbf{X}^{(L)}$ and clauses $\textbf{C}^{(L)}$ where $L$ is the number of layers; and fully connected prediction network $\operatorname{FC}(\cdot)$ maps node features to their labels.

To pre-train the three parts, we build a supervised loss function to update the parameters. We leverage the binary cross entropy (BCE) as the objective of classification:
\begin{equation}
\mathcal{L}_{c}=\sum\limits_{i=1}^{n} \operatorname{BCE}(p_i,y_i),\label{lc}
\end{equation}
where $y_i\in\{0,1\}$ is the label that describes the truth assignment of the variable generated from the MAXHS solver, and $p_i\in\{0,1\}$ is the prediction of the pre-training model from the prediction network $\operatorname{FC}(\cdot)$. It is worth noting that we only consider the classification loss under $n$ variable nodes. Since the truth assignments of clauses are easily affected by variables, it is difficult to determine the labels of clauses. Therefore, to avoid the disturbance brought by inaccurate labeling, we only consider the classification of variables.

\subsubsection{Fine-Tuning with Domain Adaptation}

After the pre-training process, we further leverage Max-SAT to build a fine-tuning process based on domain adaptation. In general, domain adaptation aims to learn domain-invariant features that can be used for different tasks. For a target CO problem, the proposed adaptation network treats Max-SAT as the source domain and learns transferable features by combining the samples from the CO and Max-SAT problems. Before the feature extraction, we can construct the data from source and target domains via transformation operations in section \ref{tranfer_G}. The data for the source domain is generated directly from Max-SAT with different distributions following the pre-training process while the data of the target domain is generated from Max-SAT that is converted from a certain CO problem.   

The network architecture of fine-tuning is based on the pre-trained network. The pre-training network learns generalizable features while the fine-tuning network focuses on learning task-specific features. With the help of pre-training, the model can converge faster and achieve better generalization performance. Concretely, the fine-tuning network contains three parts: the pre-trained feature extraction backbones based on MLP and $\operatorname{Bip-GNN(\cdot)}$ are used to extract the features from both source and target domains, the prediction or classification layers are used to predict the labels of the source and target domains, and the discriminator network $\operatorname{Dis}(\cdot)$ is to classify the domain labels for each sample.

The source domain and target domain share the same feature extraction backbone but have different classification networks. The domain adaptation framework in our work follows the supervised setting where data from both source and target domains contains labels and can be used for training. Our work can also be extended to the unsupervised setting where the labels for the target domain are unavailable. Based on the three parts, the overall loss for domain adaptation in the fine-tuning stage can be denoted as
\begin{equation}
\mathcal{L}_{ft} = \mathcal{L}_{c}+\lambda \mathcal{L}_{d},\label{lf}
\end{equation}
where $\mathcal{L}_{c}$ and $\mathcal{L}_{d}$ are the losses for classification and discrimination, and $\lambda$ is a positive hyper-parameter that controls the weight of losses. 

The discriminative loss that classifies nodes based on their domain labels can be denoted as
\begin{equation}
\mathcal{L}_{d}=(\mathbf{X}_S, \mathbf{X}_T)=\mathbb{E}_{\mathbf{x}_S \in \mathcal{D}_S} [\log \big(1-\operatorname{Dis}(\operatorname{Bip-GNN}(\textbf{x}_S))\big)]+\mathbb{E}_{\mathbf{x}_T\in \mathcal{D}_T} [\log\big(1-\operatorname{Dis}(\operatorname{Bip-GNN}(\textbf{x}_T))\big)],
\end{equation}
where $\mathbf{X}_S$ and $\mathbf{X}_D$ are the node features from source and target domains, $\mathcal{D}_S$ and $\mathcal{D}_T$ represent the data distribution of source and target domains, $\operatorname{Dis}(\cdot)$ is the domain classifier, and $\operatorname{Bip-GNN}(\cdot)$ is the feature extractor.

\subsection{Inference with Local Search Method}

The last step of our model is to search the feasible solutions for CO problems with constraint conditions by using the predictions generated by our framework. We can also get close to optimal solutions with a search algorithm. For Max-Cut problems without hard clauses, our model can directly infer the truth assignment and obtain the target values. For problems with hard clauses, we introduce a heuristic local search method to discretely obtain feasible solutions that satisfy constraints and obtain the target values. 

Local search is a typical search method for discrete optimization. We adopt a 2-improvement local search algorithm \cite{andrade2012fast, feo1994greedy} that iterates over all nodes in the graph and attempts to replace a 1-labeled node $v_i$ with two 1-labeled nodes $v_j$ and $v_k$. For the MIS problem, $v_j$ and $v_k$ must be the neighbors of $v_i$. These two nodes are 1-tight and are not connected. Here, a node is 1-tight if exactly one of its neighbors is 1-labeled. In other words, $v_i$ is the only 1-labeled neighbor of $v_j$ and $v_k$ in the graph. Notice that this local search algorithm can find a valid 2-improvement in $O(E)$ time if it exists. We introduce this method with fixed steps to find a feasible solution and evaluate the performance of our model.

\subsection{Overall Algorithms}

Most of the previous works focus on modeling specific CO problems. In contrast, our method captures the common knowledge of COs and designs a unified learning method to utilize the knowledge to solve COs. To summarize, our framework has three essential processes.

The problem transfer process leverages Max-SAT to bridge various COs. The COs are converted to Max-SAT with a general form that can capture logical information within CO problems. From the perspective of data transformation, the graphs in COs are firstly converted into clauses and then converted to bipartite graphs. The overall algorithm for problem transfer is listed in Algorithm \ref{al1}.

\begin{algorithm}[htbp]
\caption{Problem Transfer via Max-SAT}
\label{al1}
\renewcommand{\algorithmicrequire}{\textbf{Input:}}
\renewcommand{\algorithmicensure}{\textbf{Output:}}
\begin{algorithmic}[1]
    \REQUIRE A CO graph $\mathcal{G}=(\mathcal{V},\mathcal{E})$;\\
    Objective function: $\mathop{\max} f(\mathcal{S})$;\\
    Constraint condition: $g_i(\mathcal{S})\leq  b_i, b_i\in \Omega$.    
    \ENSURE A bipartite graph $\mathcal{\tilde{G}}=(\mathcal{\tilde{V}}_x,\mathcal{\tilde{V}}_c, \mathcal{\tilde{E}})$.
    
    $\#$\textbf{From a graph to Max-SAT clauses}$\#$ 
    
    \STATE Generate soft clauses $\mathcal{C}_s$ with objective function $\mathop{\max} f(\mathcal{S})$;

    \STATE Generate hard clauses $\mathcal{C}_h$ with constraint condition $g_i(\mathcal{S})\leq  b_i, b_i\in \Omega$;
    
    $\#$\textbf{From Max-SAT clauses to a bipartite graph}$\#$
    
    \STATE Construct nodes $\mathcal{\tilde{V}}_x$ and $\mathcal{\tilde{V}}_c$ for each variable $v_i$ and clause $c_j$.
    \STATE Construct edges $e^{\prime}_{(v_i,c_j)}\in \mathcal{\tilde{E}}, \forall v_i\in c_j$.
    
\end{algorithmic} 
\end{algorithm}

The pre-training process uses samples from Max-SAT to learn generalizable features that can benefit all COs. The networks used for pre-training include the MLP, feature extraction backbones based on GNNs, and the classification network. The overall algorithm for pre-training is listed in Algorithm \ref{al2}.

The fine-tuning process uses samples from Max-SAT and target CO to build a domain adaptation architecture. Based on the pre-trained network, the fine-tuning network introduces an additional discriminator for domain classification, which can learn domain-invariant features to further improve the generalizability of features. The overall algorithm for fine-tuning is listed in Algorithm \ref{al3}.

\begin{algorithm}[!htb]
\caption{Pre-training Process}
\label{al2}
\renewcommand{\algorithmicrequire}{\textbf{Input:}}
\renewcommand{\algorithmicensure}{\textbf{Output:}}
\begin{algorithmic}[1]
    \REQUIRE A bipartite graph $\mathcal{\tilde{G}}=(\mathcal{\tilde{V}}_x,\mathcal{\tilde{V}}_c, \mathcal{\tilde{E}})$ from Max-SAT,\\
    True assignment $\mathcal{Y}$,\\
    Number of pre-training layers $L$,\\
    Number of iterations $iters$.    
    \ENSURE Pre-trained parameters $\Theta_{p}$ for MLP, Bip-GNN, and FC.
    
    \STATE Initialize node attributes $\textbf{X}_{ini}$ and $\textbf{C}_{ini}$;
    \FOR{ $iter \leq iters$}
      \STATE Obtain features $\textbf{X}^{(0)}=\operatorname{MLP}(\textbf{X}_{ini})$ and $\textbf{C}^{(0)}=\operatorname{MLP}(\textbf{C}_{ini})$ for nodes of variables and clauses;
      \FOR{ $l \leq L$}
      \STATE Extract features for clauses $\textbf{C}^{(l)}$ via clause-wise aggregation Eqs. \eqref{var_to_cla} and \eqref{var_to_cla2};
      \STATE Extract features for variables $\textbf{X}^{(l)}$ via variable-wise aggregation Eqs. \eqref{cla_to_var} and \eqref{cla_to_var2};
      \ENDFOR
      
     \STATE Map variable features to the predictions via $\operatorname{FC}(\textbf{X}^{(L)})$;

      \STATE Compute the loss $\mathcal{L}_c$ in Eq. \eqref{lc};
    
      \STATE Backward and update parameters;
    \ENDFOR
    \STATE Obtain the updated parameters $\Theta_{p}$.
    
\end{algorithmic} 
\end{algorithm}
\begin{algorithm}[!htb]
\caption{Fine-tuning Based on Domain Adaptation}
\label{al3}
\renewcommand{\algorithmicrequire}{\textbf{Input:}}
\renewcommand{\algorithmicensure}{\textbf{Output:}}
\begin{algorithmic}[1]
    \REQUIRE A bipartite graph $\mathcal{\tilde{G}}=(\mathcal{\tilde{V}}_x,\mathcal{\tilde{V}}_c, \mathcal{\tilde{E}})$ from Max-SAT,\\
    Number of fine-tuning layers $L$,\\
    Number of iterations $iters$,\\
    True assignment $\mathcal{Y}$.
    \ENSURE Predicted feasible solutions $\mathcal{P}=\{p_{i}\}_{i=1}^{n}$; \\
    Fine-tuned parameters $\Theta_{ft}$ for MLP, Bip-GNN, FC, and Dis.
        
    \STATE Initialize node attributes of both domains $\textbf{X}_{ini}$ and $\textbf{C}_{ini}$;

    \FOR{ $iter \leq iters$}
      \STATE Obtain features $\textbf{X}^{(0)}_{S}=\operatorname{MLP}(\textbf{X}_{ini})$ and $\textbf{C}^{(0)}_{S}=\operatorname{MLP}(\textbf{C}_{ini})$ for source variables and clauses;

      \STATE Obtain features $\textbf{X}^{(0)}_{T}=\operatorname{MLP}(\textbf{X}_{ini})$ and $\textbf{C}^{(0)}_{T}=\operatorname{MLP}(\textbf{C}_{ini})$ for target variables and clauses;

    \FOR{ $l \leq L $}
      \STATE Extract features for clauses $\textbf{C}^{(l)}_{S}$ and $\textbf{C}^{(l)}_{T}$ via clause-wise aggregation Eqs. \eqref{var_to_cla} and \eqref{var_to_cla2};
      \STATE Extract features for variables $\textbf{X}^{(l)}_{S}$ and $\textbf{X}^{(l)}_{T}$ via variable-wise aggregation Eqs. \eqref{cla_to_var} and \eqref{cla_to_var2};
      \ENDFOR   

    \STATE Map variable features to the assignment labels via $\operatorname{FC}(\textbf{X}^{(L)}_S)$ and $\operatorname{FC}(\textbf{X}^{(L)}_T)$;

    \STATE Map variable features to the domain labels via $\operatorname{Dis}(\textbf{X}^{(L)}_S)$ and $\operatorname{Dis}(\textbf{X}^{(L)}_T)$;
  
    \STATE Compute the loss $\mathcal{L}_{ft}$ in Eq. \eqref{lf};
    
    \STATE Updated parameters $\Theta_{ft}$ for MLP, Bip-GNN, Dis, and FC;
    
    \ENDFOR

    \STATE Obtain the optimal parameters $\Theta_{ft}$;
    \STATE Predict feasible solutions $\mathcal{P}$.
    
\end{algorithmic} 
\end{algorithm}

\section{Experimental Results}
In this section, we evaluated the effectiveness of the proposed framework for three CO problems on graphs, i.e., Max-Cut, MIS, and MDS. Generally, we aimed to answer three essential questions by the experiments. \textbf{Q1}: Can we leverage Max-SAT to learn transferable and generalizable features that can improve the ability to solve different CO problems on graphs? \textbf{Q2}: How can we incorporate Max-SAT into the pipelines of GNN-based learning frameworks to fully utilize the information of Max-SAT? \textbf{Q3}: Do different pre-training and domain adaptation strategies matter for the ability of solving COs?

\subsection{Experimental Setup}
\paragraph{Datasets}
We introduce the datasets that were used in our experiments in this part. In the pre-training stage, a large number of Max-SAT instances were required. We generated the clauses by running three representative generators with different distributions: the uniform distribution, the power-law distribution, and the double power-law distribution. A total number of 20,000 Max-SAT instances were generated and the MaxHS solver was used to obtain the solutions that served as labels for training. In the fine-tuning stage, according to the CO problems, different datasets are leveraged. For Max-Cut, we introduce the GSET benchmark\footnote{http://web.stanford.edu/~yyye/yyye/Gset/} for evaluation. GSET is a set of 71 unweighted graphs that were commonly used for testing the algorithms to solve the Max-Cut problem. For MIS and MDS problems, we introduce the frb benchmark with four different instance scales for evaluation. Each frb dataset consists of five instances of hard CO problems and is regarded as the benchmark for CSP competition annually.

\paragraph{Evaluation Metrics}
For different CO problems, we used the corresponding evaluation metrics based on their feasible solutions. For Max-Cut, $p$ values \cite{dembo2017extremal} were used to evaluate the number of graph cuts. Concretely, $p$ can be calculated by
\begin{equation}
p(z)=\frac{z/n-\gamma/4}{\sqrt{\gamma/4}},
\end{equation}
where $z$ is the predicted cut size for a $\gamma$-regular graph with $n$ nodes and $\gamma$ is the degree of nodes. For MIS and MDS, the number of cuts or sets is used for evaluation.

\paragraph{Baseline methods} To evaluate our method, we compare it with several baseline methods that have been widely used in solving COs. These methods can be roughly classified as traditional methods, heuristic methods, and learning-based methods. \textbf{SDP} \cite{choi2000solving} is a classical approach based on semi-definite programming. \textbf{EO} \cite{boettcher2001extremal} is the extremal optimization method. \textbf{BLS} \cite{benlic2013breakout} is the breakout local search method. \textbf{ECO-DQN} \cite{barrett2020exploratory} is an approach of exploratory CO with reinforcement learning. \textbf{GMC-A} and \textbf{GMC-B} \cite{yao2019experimental} are two unsupervised GNN architectures for Max-CUT with different loss functions. \textbf{RUN-CSP} \cite{toenshoff2021graph} is a recurrent unsupervised neural network for constraint satisfaction problems. \textbf{PIGNN} \cite{schuetz2022combinatorial} is the physics-inspired GNNs for CO problems. \textbf{MAXSAT} \cite{liu2023can} is a GNN-based framework designed for Max-SAT problems.

\paragraph{Experimental Settings}
The proposed framework was implemented in Python with PyTorch. We summarized the hyper-parameters used in our paper as follows. The dimension of features $d$ was set as 128. The number of GNN layers $L$ for pre-training and fine-tuning was set as 5 and the first 2 layers were fixed during the fine-tuning. The number of layers for MLP was set as 2. The number of epochs for pre-training and fine-tuning were tuned and set as 400 and 100. The Adam optimizer was used for model training with a learning rate of $10^{-5}$ and a weight decay of $10^{-10}$. The warm-up trick was also adopted. The number of FC layers for classification was set as 2 and the discriminator used a two-layer fully connected network to generate domain labels. The steps for local search for MIS and MDS were set as 120. All experiments were conducted on a workstation equipped with an Intel Xeon(R) Gold 6139M CPU 515 @ 2.30GHz and an NVIDIA RTX 3090 GPU with 24GB.

\begin{table}[!htbp]
\caption{Results of solving Max-Cut with different approaches on random graphs.}
\label{Max-Cut-Random}
\resizebox{\textwidth}{!}{
\centering
\begin{tabular}{cccccccccccccccc}
\toprule
\multirow{2}{*}{Methods} & \multicolumn{3}{c}{n=100} & \multicolumn{3}{c}{n=200}&
\multicolumn{3}{c}{n=500} & \multicolumn{3}{c}{n=800} & \multicolumn{3}{c}{n=1000} \\
&$\gamma=3$ & $\gamma=5$ & $\gamma=10$ & $\gamma=3$ & $\gamma=5$ & $\gamma=10$ & $\gamma=3$ & $\gamma=5$ & $\gamma=10$ & $\gamma=3$ & $\gamma=5$ & $\gamma=10$ & $\gamma=3$ & $\gamma=5$ & $\gamma=10$ \\
\hline
SDP & 0.709 & 0.697 & 0.689 & 0.709 & 0.702 & 0.692 & 0.702 & 0.690 & 0.682 & 0.701 & 0.688 & 0.679 & - & - & - \\
EO & 0.712 & \textbf{0.708} & 0.703 & 0.721 & \textbf{0.723} & 0.724 & 0.727 & 0.737 & \textbf{0.735} & - & - & - & - & - & - \\
\hline
BLS & 0.712 & 0.707 & 0.704 & 0.720 & 0.721 & 0.719 & 0.722 & 0.725 & 0.721 & 0.720 & 0.726 & 0.717 & 0.721 & 0.725 & 0.720 \\
ECO-DQN & \textbf{0.713} & 0.705 & 0.707 & 0.718 & \textbf{0.723} & 0.720 & 0.725 & 0.727 & 0.725 & 0.722 & 0.721 & 0.722 & 0.726 & 0.726 & 0.721 \\
\hline
GMC-A &0.691&0.655&0.637&0.701&0.683&0.660&0.693&0.668&0.599&0.691&0.666&0.602&0.688&0.662&0.601 \\
GMC-B &0.698&6.660&0.630&0.708&0.667&0.646&0.707&0.701&0.670&0.699&0.696&0.653&0.702&0.694&0.651 \\
RUN-CSP &0.702&0.706&0.704&0.711&0.715&0.714&0.714&0.726&0.710&0.704&0.713&0.705&0.705&0.711&0.702 \\
PI-GNN &0.704&0.706&\textbf{0.708}&0.712&0.718&0.717&0.715&0.726&0.716&0.711&0.724&0.712&0.709&0.722&0.712 \\
MAX-SAT &0.702&0.705&\textbf{0.708}&0.712&0.716&0.718&0.721&0.732&0.726&0.719&0.730&0.728&0.718&0.730&0.726 \\
Ours &0.702 & \textbf{0.708}&0.707&\textbf{0.722}&0.721&\textbf{0.725}&\textbf{0.732}&\textbf{0.738}&0.733 & \textbf{0.734}&\textbf{0.739}&\textbf{0.733}&\textbf{0.732}&\textbf{0.738}&\textbf{0.733} \\
\bottomrule
\end{tabular}
}
\end{table}

\begin{table}[!htb]  
\centering  
\begin{threeparttable}  
\caption{Results of solving Max-Cut with different approaches on the GSET benchmark.}		
\begin{tabular}{ccccccc}  
\toprule  
\multicolumn{5}{c}{Max-Cut}&\cr
\midrule
\multicolumn{1}{c}{Methods}&  
\multicolumn{1}{c}{G14}&
\multicolumn{1}{c}{G15}&
\multicolumn{1}{c}{G22}&
\multicolumn{1}{c}{G55}\cr  
\midrule  
$V$ &800&800&2000&5000\cr  
$E$ &4694&4661&19990&12468\cr   
\midrule
SDP &2922&2938&-&-\cr
EO &2991&3047&-&-\cr
BLS &\textbf{3064}&\textbf{3050}&-&-\cr			
ECO-DQN &3056&3028&13268&10198\cr
\midrule
GMC-A &2940&2948&12978&10094\cr
GMC-B &2918&2906&13246&10104\cr
RUN-CSP &2943&2928&13028&10116\cr
PI-GNN &3026&2990&13181&10138\cr
MAX-SAT &3046&3012&13250&10204\cr
Ours &3052&3018&\textbf{13269}&\textbf{10208}&\cr
\bottomrule  
\end{tabular}            
\label{Max-Cut-GSET}
\end{threeparttable}  
\end{table}

\begin{table}[!htb]  
\centering  
\caption{Results of solving MIS with different approaches on frb datasets.}
\label{MIS}
\begin{threeparttable}   
\begin{tabular}{ccccccc}  
\toprule  
\multicolumn{5}{c}{Maximum Independent Set}&\cr
\midrule
\multicolumn{1}{c}{Methods}&  
\multicolumn{1}{c}{frb30–15}&
\multicolumn{1}{c}{frb40–19}&
\multicolumn{1}{c}{frb50–23}&
\multicolumn{1}{c}{frb59–26}\cr  
\midrule  
V&450&760&1150&1534\cr  
E&18k&41k&80k&126k\cr
\midrule 
ReduMIS(Solver) &30.0&39.4&48.8&57.4\cr
\midrule 
Greedy &24.6&33.0&42.2&48.0\cr
\midrule 
GMC-A &21.2&27.7&40.2&45.2\cr
GMC-B &20.8&28.1&40.6&46.0\cr
RUN-CSP &25.8&33.6&42.2&49.4\cr
PIGNN &26.5&35.1&44.4&51.8\cr
MAX-SAT &\textbf{27.9}&36.6&44.4&51.8\cr
Ours &27.8&\textbf{36.8}&\textbf{44.8}&\textbf{52.2}&\cr
\bottomrule  
\end{tabular}  
\end{threeparttable}  
\end{table}

\begin{table}[!htb]  
\centering  
\begin{threeparttable}  
\caption{Results of solving MDS with different approaches on frb datasets.} 
\label{MDS}  
\begin{tabular}{ccccccc}
\toprule  
\multicolumn{5}{c}{Minimum Dominated Set}&\cr
\midrule
\multicolumn{1}{c}{Methods}&  
\multicolumn{1}{c}{frb30–15}&
\multicolumn{1}{c}{frb40–19}&
\multicolumn{1}{c}{frb50–23}&
\multicolumn{1}{c}{frb59–26}\cr  
\midrule  
V&450&760&1150&1534\cr  
E&18k&41k&80k&126k\cr
\midrule 
Greedy&191.1&\textbf{255.0}&\textbf{296.5}&284.8\cr
HTS-DS&191.0&255.2&297.6&286.4\cr
MWDS-CRO&191.0&255.1&298.4&285.5\cr
\midrule
GMC-A&201.2&268.1&301.3&292.2\cr
GMC-B&192.1&261.4&299.1&291.6\cr
RUN-CSP&\textbf{190.8}&255.4&299.3&278.6\cr
PIGNN&191.1&255.3&299.1&280.2\cr
MAX-SAT&191.0&255.3&298.8&278.8&\cr
Ours&\textbf{190.8}&255.1&298.5&\textbf{278.1}&\cr
\bottomrule  
\end{tabular}  
\end{threeparttable}  
\end{table}

\subsection{Performance of Solving COs}

In this section, we evaluated the effectiveness of the proposed framework and aimed to answer \textbf{Q1}. Without loss of generality, Max-Cut, MIS, and MDS are selected for evaluation. Our framework is also adaptive to other COs that can be transformed as Max-SAT. 

For Max-Cut, we first reported the mean $p$-values across 1,000 regular graphs with 100, 200, 500, and 1000 nodes for three levels of node degrees in Table \ref{Max-Cut-Random}. Then, the experiments on the GSET benchmark were carried out and the number of cuts was reported in Table \ref{Max-Cut-GSET} for comparison. When encountering large-scale datasets (G22 and G55), traditional methods, e.g., EO and BLS, were out of memory and could not obtain the results. Based on the two tables, we have the following observations: (1) Our method obtained the best results compared with learning-based methods, which demonstrated that the proposed framework exhibited the powerful ability to learn useful information from CO problems. The results also showed that Max-SAT can boost the performance of solving COs; (2) Our method was superior in most cases compared with all baselines. The advantages of our method were mainly in large-scale Max-Cut problems (e.g., $n$=800, G22, and G55) where the performance surpassed traditional methods; (3) Traditional or heuristic methods are very competitive in Max-Cut. However, the gaps between our method and these methods were small. Besides, the time of training and inference for our method was much less than these methods. 

For MIS and MDS, we reported the number of sets for MIS and the number of dominated nodes for MDS in Tables \ref{MIS} and \ref{MDS}. $V$ and $E$ are the sets for nodes and edges. Greedy denotes the classical greedy algorithm. Larger and smaller values denote better results for MIS and MDS, respectively. The experiments were carried out on four frb datasets. Similar results can be observed in these two CO problems. Particularly, In MIS, our results were relatively closer to the results obtained by the solver, which showed the effectiveness of our framework.

Based on the results obtained on the above three CO problems, we can answer \textbf{Q1}: Max-SAT offered common knowledge and the pre-training and adaptation pipeline built based on Max-SAT can learn transferable and generalizable features to improve the ability to solve different CO problems on graphs.

\begin{figure}[!htb]
\centering
\includegraphics[width=\linewidth]{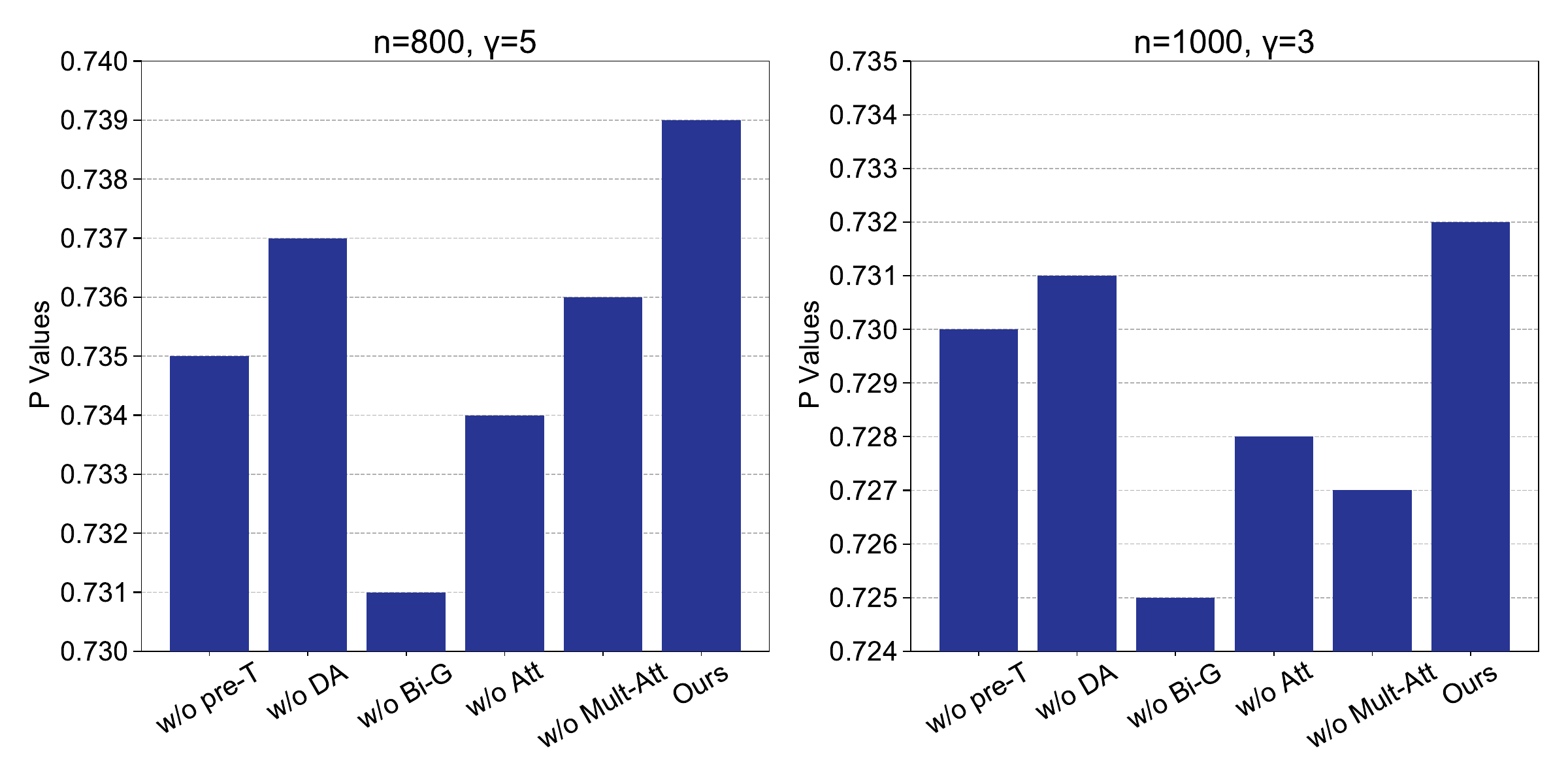}
\caption{Results of ablation study with different variants of our model on Max-Cut.}
\label{ablation_study}
\end{figure}

\begin{figure}[!htb]
\centering
\includegraphics[width=0.9\linewidth]{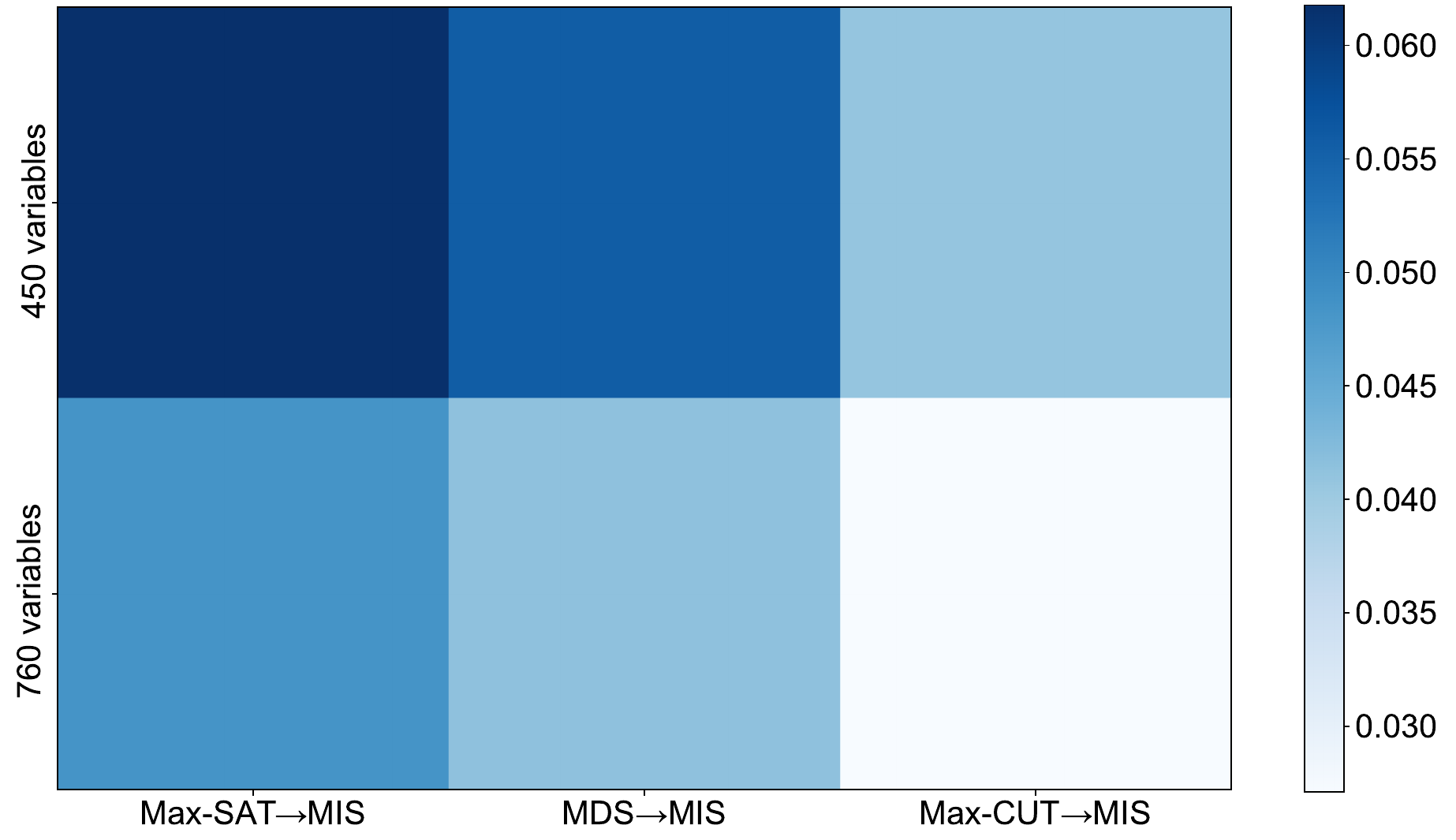}
\caption{Heat maps of knowledge transferability from three different CO problems to the maximum independent set problem on two scales of nodes.}
\label{heat_map}
\end{figure}

\subsection{Ablation Study}
In this part, we conducted ablation studies on the Max-Cut problem to verify the effectiveness of each module in our framework. We gradually removed each single module and kept the other modules unchanged. The variants of our frameworks are listed as follows. \textbf{w/o pre-T} denotes the model that removes the pre-training stage but keeps the domain adaptation module. \textbf{w/o DA} denotes the model that removes the domain adaptation module but keeps the pre-training stage.
\textbf{w/o Bi-G} is the model that directly uses GNNs to extract the features in the original graphs represented from CO problems without converting them to MAX-SAT and bipartite graphs. \textbf{w/o Att} is the model that keeps bipartite GNNs without the attention mechanism. \textbf{w/o Mult-Att} is the model that keeps bipartite GNNs without the multi-head mechanism.

We reported the $p$-values for all variants in Figure \ref{ablation_study} and had the following observations. (1) All modules boosted the performance and played important roles in our framework. The overall model obtained the best results, which demonstrated that these modules were nicely incorporated into our framework. (2) Removing the problem transfer step experienced the largest performance degradation, which proved the effectiveness of Max-SAT and also answered \textbf{Q1}. (3) Pre-training and adaptation were both useful in improving the performance. Since both steps utilized Max-SAT to assist the learning of transferable features, we can answer \textbf{Q2} and conclude that the combination of different strategies to use Max-SAT can largely benefit solving CO problems. (4) Attention and multi-head mechanisms also affected the results, which showed that better feature extraction backbones indeed helped the learning methods to solve CO problems.

\begin{figure}[!htb]
\centering
\includegraphics[width=\linewidth]{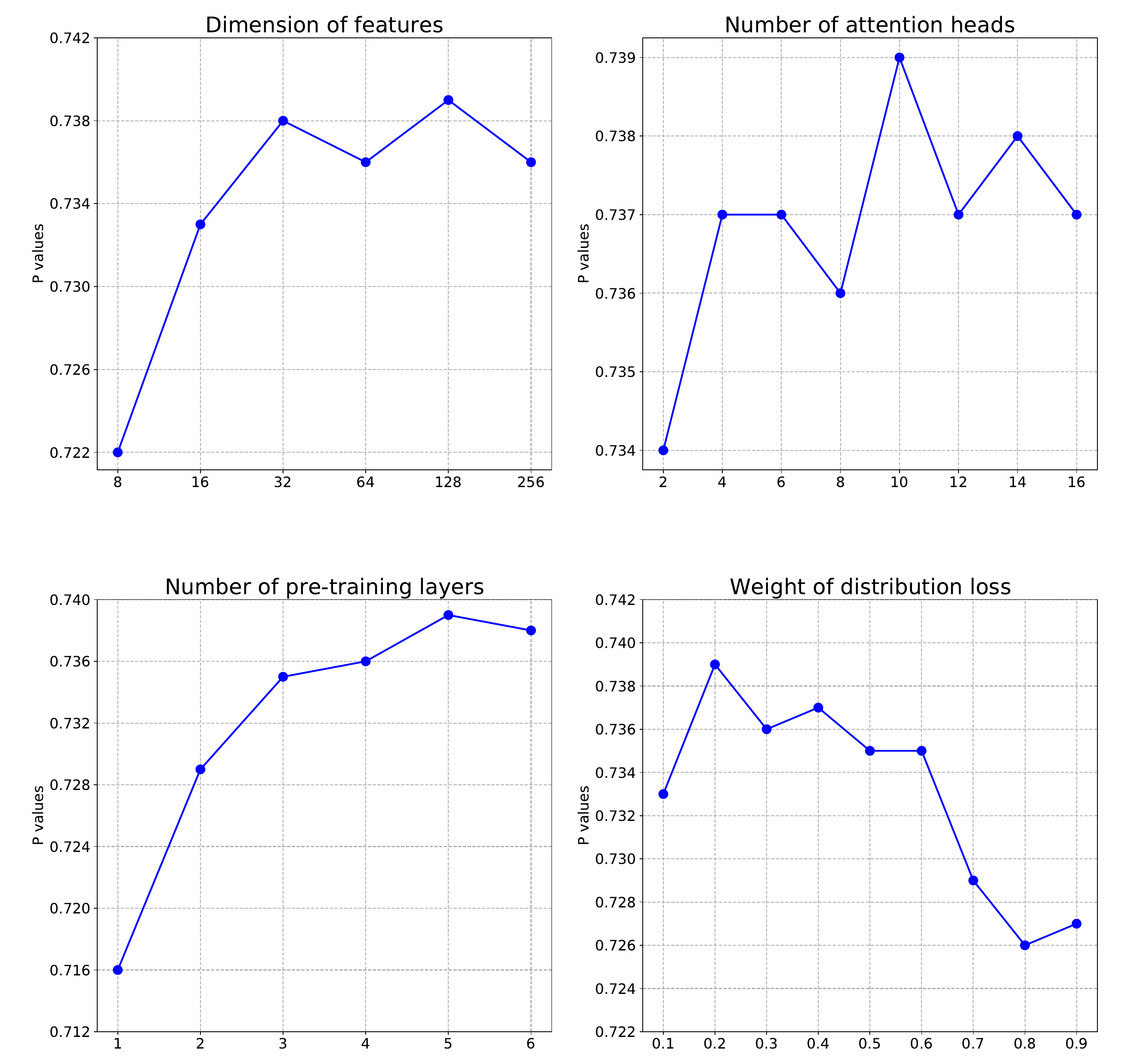}
\caption{Parameter analysis for feature dimension $d$, number of attention heads $h$, weight of loss $\lambda$, and number of layers $L$.}
\label{parameter}
\end{figure}

\subsection{Effects of Different Knowledge Transfer Strategies}

In this subsection, we verify the knowledge transfer ability of our model. we fixed the target domain as the MIS problem and used Max-SAT, Max-Cut, and MDS as source domains to check different degrees of transferability. To be concrete, we fed the data from different source domains into our framework for pre-training and domain adaptation and observed how the performance was affected by the selection of domains. We selected two scales of nodes, i.e., 450 variables and 760 variables, for demonstration. The results were shown in Figure \ref{heat_map} where darker colors indicated better performance and transferability.

From the heat maps, we can observe that Max-SAT$\rightarrow$MIS always obtained the best results compared with the other two types of knowledge transfer, which showed that Max-SAT was more general and had the strongest transferability across different CO problems. Moreover, MDS is a more relevant CO problem to MIS compared with Max-Cut. Therefore, we observed better results from MDS$\rightarrow$MIS than Max-Cut$\rightarrow$MIS. This observation answered \textbf{Q3} where different source domains do affect the performance of the target domain. From a practical viewpoint, selecting an appropriate source domain was always hard, and irrelevant tasks would generate negative transfers and eventually affect the performance of the target domain. 

\subsection{Parameter Analysis}
In this subsection, we analyzed the influence of several significant hyper-parameters on the model performance: the dimension of features $d$, the number of attention heads $h$, the weight of losses $\lambda$, and the number of pertaining layers $L$. We recorded the performance of different $d$, $h$, $\lambda$, $L$ values on Max-CUT. The results were reported in Figure \ref{parameter}. It can be observed that with different dimensions of features, our model exhibited relatively stable performance when $d$ was greater than or equal to 16. Our model achieved optimal performance when $d$ was 128. Our model was not significantly affected by the number of attention heads. It achieved optimal performance when $k$ was 10. The optimal value of weight $\lambda$ was obtained in 0.2. Pre-training led to better results, especially when the number of layers exceeded three.

\section{Conclusion}
In this study, we aimed to discover the transferability of deep learning architectures for solving different CO problems on graphs. To achieve this goal, we leveraged the Max-SAT problem as a general formulation of original CO problems. The key advantages were two-fold. First, Max-SAT served as a bridge for various CO problems and provided a tool for discovering common properties in CO problems. Second, Max-SAT problems were represented as clauses that included logical information in addition to the simple correlations in graphs. By further constructing a pre-training and adaptation framework, we can extract transferable features that can be used for various CO problems. Experiments showed that incorporating Max-SAT indeed improved the ability to solve various CO problems on graphs.

Our current work can be extended from the following directions. The pre-training stage is built on supervised tasks where labeled data samples are required. As a future direction, it is necessary to develop unsupervised pre-training paradigms that can fully leverage information from the data itself. Moreover, advanced GNNs that are more suitable for the properties of CO problems should be designed in the future.

\paragraph{Acknowledgements} This work was supported by the National Natural Science Foundation of China (Grant No. 11991021, No. 11991020, No. 12271503).

\bibliographystyle{unsrt}
\bibliography{re}
\end{document}